\documentclass[11pt,a4paper]{article}
\usepackage{xr-hyper}
\usepackage[hyperref]{acl2021}
\usepackage{times}
\usepackage{latexsym}
\usepackage{bbding,pifont,adjustbox,amssymb} 
\usepackage{url}

\usepackage{microtype}

\usepackage[T1]{fontenc}
\usepackage[scaled=.8]{beramono}
\usepackage[scaled=.95]{helvet}
\usepackage{amsmath}
\usepackage{color,soul}

\usepackage{subcaption}

\usepackage[shortlabels]{enumitem} 
\setitemize{noitemsep,topsep=0em} 

\usepackage{multirow}
\usepackage{tikz}
\usepackage{tikz-dependency}
\usepackage[warn]{textcomp}
\usepackage{etoolbox}
\usepackage{xfrac}
\usepackage{pgfplotstable}

\usepackage{tabularx}
\usepackage{collcell}
\usepackage{booktabs}
\usepackage{nameref}
\usepackage{cleveref}

\pgfplotsset{compat=1.14}

\usepackage{wrapfig}

\makeatletter
\newcommand*{\addFileDependency}[1]{
  \typeout{(#1)}
  \@addtofilelist{#1}
  \IfFileExists{#1}{}{\typeout{No file #1.}}
}
\makeatother

\newcommand*{\myexternaldocument}[1]{%
    \externaldocument{#1}%
    \addFileDependency{#1.tex}%
    \addFileDependency{#1.aux}%
}

\myexternaldocument{implicit-parser-supp}

\aclfinalcopy 

\title{Meaning Representation of Numeric Fused-Heads in UCCA}

\author{
  Ruixiang Cui \and Daniel Hershcovich \\
  Department of Computer Science \\
  University of Copenhagen \\
  \quad\texttt{\{rc, dh\}@di.ku.dk} \\
}

\date{}

\begin{document}
\maketitle

\section{Meaning Representations}
Despite the usefulness and popularity of syntactic representation frameworks such as Universal Dependencies \citep{nivre-etal-2020-universal}, researchers have pointed out their sole focus on surface-syntactic information \citep{hershcovich2019content, zabokrtsky-etal-2020-sentence} and proposed to design linguistic schemes on a deep-syntactic or semantic level. Meaning representation (MR) frameworks aim at capturing shared semantic principles in one or more languages and representing sentences in a structured way \citep{zabokrtsky-etal-2020-sentence}. Frameworks including as Elementary Dependency Structures \citep[EDS;][]{oepen2016towards}, Prague Tectogrammatical Graphs \citep[PTG;][]{hajic-etal-2020-prague}, Universal Conceptual Cognitive Annotation \citep[UCCA;][]{abend2013universal} and Abstract Meaning Representation \citep[AMR;][]{banarescu2013abstract} are designed to represent semantics directly and (to varying degrees) comprehensively. Whereas these frameworks model explicit linguistic constructions, only few address implicit phenomena, where predicates, arguments and modifiers are omitted and need to be inferred from the context. One of them is UCCA, which annotates implicit units as part of the meaning representation graph. Furthermore, \citet{cui-hershcovich-2020-refining} proposed refining this annotation with a layer representing fine-grained distinctions between implicit arguments that correspond to participants.

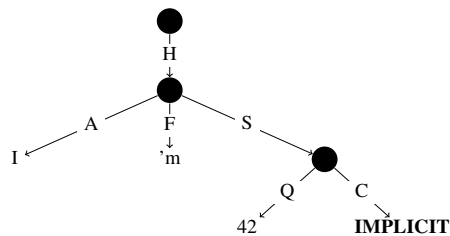
\begin{figure}[t]
\centering
\resizebox{6cm}{!}{%
\begin{tikzpicture}[->,level distance=1.35cm,
  level 1/.style={sibling distance=4cm},
  level 2/.style={sibling distance=30mm},
  level 3/.style={sibling distance=30mm},
  every circle node/.append style={fill=black},
  every node/.append style={text height=1ex,text depth=0}]
  \tikzstyle{word} = [font=\rmfamily,color=black]
  \node (1_1) [circle] {}
  {
  child {node (1_2) [circle] {}
    {
    child {node (1_3) [word] {I}  edge from parent node[midway, fill=white]  {A}}
    child {node (1_4) [word] {'m}  edge from parent node[midway, fill=white]  {F}}
    child {node (1_5) [circle] {}     
      {
      child {node (1_6) [word] {42}  edge from parent node[midway, fill=white]  {Q}}
      child {node (1_11) [word] {\textbf{IMPLICIT}}  edge from parent node[midway, fill=white]  {C}}
      } edge from parent node[midway, fill=white]  {S}}
    } edge from parent node[midway, fill=white]  {H}
    }
  };
\end{tikzpicture}
}
\caption{UCCA example of a sentence with NFH.}
\label{fig:example 1}
\end{figure}


\begin{table*}[t]
\centering
\begin{adjustbox}{width=375pt,margin=1pt}
    \begin{tabular}{l|llllllllllllll}
    & A & C & D & E & H & L & P & Q & R & S & T & S/A & $\emptyset$ & Total \\ \hline
    w/ NFH & 24 & 80 & 2 & 6 & 1 & 0 & 4 & 68 & 5 & 8 & 3 & 1 & 2 & 206 \\
    w/o NFH & 5 & 22 & 6 & 6 & 0 & 1 & 4 & 240 & 3 & 3 & 0 & 0 & 3 & 294
    \end{tabular}
\end{adjustbox}
    \caption{Parser predictions for the NFH test set. The upper row represents instances supposed to have numeric fused-heads. The lower row are instances without numeric fused-heads. The parser predicted Quantifier (Q) for 33\% of the NFH exmaples in the test set, and 81\% of the \textit{non-}NFH examples in the test set. Empty sets are failed predictions that cannot be converted from MRP format to XML for measurements.}
    \label{tab:results}
\end{table*}

\section{Numeric Fused-Heads}
The linguistic construction of fused heads \citep[FHs;][]{elazar-goldberg-2019-wheres} corresponds to a noun phrase where the head noun is ``fused'' with the dependent modifier. Numeric fused-heads (NFHs), where the phrase is a number, are an interesting subset. For example, in the sentence `I'm 42' (Figure~\ref{fig:example 1}), the phrase `42' refers to the age of the person in years, though \textit{years} is implicit in this context. We investigate whether the semantics of this construction is represented in meaning representation frameworks in English, empirically with UCCA.

\section{NFHs in UCCA}
For UCCA, the annotation guidelines \cite{abend2020uccas} give only a very partial account of NFHs, and it is not clear how implicit NFHs should be annotated, for example. However, our interpretation of the guidelines would assign this construction the UCCA annotation of a Quantifier (Q) with an implicit or remote Center (C), as in Figure~\ref{fig:example 1}. We inspect the UCCA English Web Treebank Corpus (UCCA EWT)\footnote{\href{https://github.com/UniversalConceptualCognitiveAnnotation/UCCA_English-EWT}{\tiny\texttt{github.com/UniversalConceptualCognitiveAnnotation/UCCA\_English-EWT}}} and the UCCA Wikipedia Corpus (UCCA Wiki)\footnote{\href{https://github.com/UniversalConceptualCognitiveAnnotation/UCCA_English-Wiki}{\tiny\texttt{github.com/UniversalConceptualCognitiveAnnotation/UCCA\_English-Wiki}}}, which are both annotated with implicit units, and find 54 and 124 instances according to this rule, respectively. There may be other instances of this linguistic phenomenon in the corpora, which we are in the progress of manually identifying.

\section{Parsing Experiments}
Subsequently, we train a recently proposed transition-based neural parser, which is able to handle implicit arguments \cite{cui2021great}, on the UCCA EWT dataset, and use its predictions on the NFH Identification test set introduced by \citet{elazar-goldberg-2019-wheres}, to extract NFH instances in a zero-shot prediction experiment, without training on the associated training set. We find that the parser failed to predict any implicit \textit{or} remote unit for any of the 206 NFH test instances in this setting. Table~\ref{tab:results} shows the results. The low performance is likely to result from the small number of examples for this phenomenon in the training set, and possibly the fact that NFHs, in fact, appear in several distinct constructions, idiosyncratically annotated in different ways in UCCA. 

\section{Conclusion}
We conclude that the implicit UCCA parser does not address NFHs consistently, which could result either from inconsistent annotation, insufficient training data or a modelling limitation. We are investigating which factors are involved. We consider this phenomenon important, as it is pervasive in text and critical for correct inference. Careful design and fine-grained annotation of NFHs in MRs would benefit downstream tasks such as machine translation, natural language inference and question answering, particularly when they require numeric reasoning \cite[e.g.,][]{dua-etal-2019-drop}, as recovering and categorizing them. We are investigating the treatment of this phenomenon by other meaning representations, such as AMR. We encourage researchers in meaning representations, and computational linguistics in general, to address this phenomenon in future research.

\bibliography{references,anthology}

\begin{thebibliography}{12}
\expandafter\ifx\csname natexlab\endcsname\relax\def\natexlab#1{#1}\fi

\bibitem[{Abend and Rappoport(2013)}]{abend2013universal}
Omri Abend and Ari Rappoport. 2013.
\newblock \href {https://www.aclweb.org/anthology/P13-1023} {{U}niversal
  {C}onceptual {C}ognitive {A}nnotation ({UCCA})}.
\newblock In \emph{Proceedings of the 51st Annual Meeting of the Association
  for Computational Linguistics (Volume 1: Long Papers)}, pages 228--238,
  Sofia, Bulgaria. Association for Computational Linguistics.

\bibitem[{Abend et~al.(2020)Abend, Schneider, Dvir, Prange, and
  Rappoport}]{abend2020uccas}
Omri Abend, Nathan Schneider, Dotan Dvir, Jakob Prange, and Ari Rappoport.
  2020.
\newblock \href {http://arxiv.org/abs/2012.15810} {Ucca's foundational layer:
  Annotation guidelines v2.1}.

\bibitem[{Banarescu et~al.(2013)Banarescu, Bonial, Cai, Georgescu, Griffitt,
  Hermjakob, Knight, Koehn, Palmer, and Schneider}]{banarescu2013abstract}
Laura Banarescu, Claire Bonial, Shu Cai, Madalina Georgescu, Kira Griffitt, Ulf
  Hermjakob, Kevin Knight, Philipp Koehn, Martha Palmer, and Nathan Schneider.
  2013.
\newblock \href {https://www.aclweb.org/anthology/W13-2322} {{A}bstract
  {M}eaning {R}epresentation for sembanking}.
\newblock In \emph{Proceedings of the 7th Linguistic Annotation Workshop and
  Interoperability with Discourse}, pages 178--186, Sofia, Bulgaria.
  Association for Computational Linguistics.

\bibitem[{Cui and Hershcovich(2020)}]{cui-hershcovich-2020-refining}
Ruixiang Cui and Daniel Hershcovich. 2020.
\newblock \href {https://www.aclweb.org/anthology/2020.dmr-1.5} {Refining
  implicit argument annotation for {UCCA}}.
\newblock In \emph{Proceedings of the Second International Workshop on
  Designing Meaning Representations}, pages 41--52, Barcelona Spain (online).
  Association for Computational Linguistics.

\bibitem[{Cui and Hershcovich(2021)}]{cui2021great}
Ruixiang Cui and Daniel Hershcovich. 2021.
\newblock Great service! fine-grained parsing of implicit arguments.
\newblock In \emph{Proceedings of the 17th International Conference on Parsing
  Technologies}.

\bibitem[{Dua et~al.(2019)Dua, Wang, Dasigi, Stanovsky, Singh, and
  Gardner}]{dua-etal-2019-drop}
Dheeru Dua, Yizhong Wang, Pradeep Dasigi, Gabriel Stanovsky, Sameer Singh, and
  Matt Gardner. 2019.
\newblock \href {https://doi.org/10.18653/v1/N19-1246} {{DROP}: A reading
  comprehension benchmark requiring discrete reasoning over paragraphs}.
\newblock In \emph{Proceedings of the 2019 Conference of the North {A}merican
  Chapter of the Association for Computational Linguistics: Human Language
  Technologies, Volume 1 (Long and Short Papers)}, pages 2368--2378,
  Minneapolis, Minnesota. Association for Computational Linguistics.

\bibitem[{Elazar and Goldberg(2019)}]{elazar-goldberg-2019-wheres}
Yanai Elazar and Yoav Goldberg. 2019.
\newblock \href {https://doi.org/10.1162/tacl_a_00280} {Where{'}s my head?
  {D}efinition, data set, and models for numeric fused-head identification and
  resolution}.
\newblock \emph{Transactions of the Association for Computational Linguistics},
  7:519--535.

\bibitem[{Haji{\v{c}} et~al.(2020)Haji{\v{c}}, Bej{\v{c}}ek, Hlavacova,
  Mikulov{\'a}, Straka, {\v{S}}t{\v{e}}p{\'a}nek, and
  {\v{S}}t{\v{e}}p{\'a}nkov{\'a}}]{hajic-etal-2020-prague}
Jan Haji{\v{c}}, Eduard Bej{\v{c}}ek, Jaroslava Hlavacova, Marie Mikulov{\'a},
  Milan Straka, Jan {\v{S}}t{\v{e}}p{\'a}nek, and Barbora
  {\v{S}}t{\v{e}}p{\'a}nkov{\'a}. 2020.
\newblock \href {https://www.aclweb.org/anthology/2020.lrec-1.641} {{P}rague
  dependency treebank - consolidated 1.0}.
\newblock In \emph{Proceedings of the 12th Language Resources and Evaluation
  Conference}, pages 5208--5218, Marseille, France. European Language Resources
  Association.

\bibitem[{Hershcovich et~al.(2019)Hershcovich, Abend, and
  Rappoport}]{hershcovich2019content}
Daniel Hershcovich, Omri Abend, and Ari Rappoport. 2019.
\newblock \href {https://doi.org/10.18653/v1/N19-1047} {Content differences in
  syntactic and semantic representation}.
\newblock In \emph{Proceedings of the 2019 Conference of the North {A}merican
  Chapter of the Association for Computational Linguistics: Human Language
  Technologies, Volume 1 (Long and Short Papers)}, pages 478--488, Minneapolis,
  Minnesota. Association for Computational Linguistics.

\bibitem[{Nivre et~al.(2020)Nivre, de~Marneffe, Ginter, Haji{\v{c}}, Manning,
  Pyysalo, Schuster, Tyers, and Zeman}]{nivre-etal-2020-universal}
Joakim Nivre, Marie-Catherine de~Marneffe, Filip Ginter, Jan Haji{\v{c}},
  Christopher~D. Manning, Sampo Pyysalo, Sebastian Schuster, Francis Tyers, and
  Daniel Zeman. 2020.
\newblock \href {https://www.aclweb.org/anthology/2020.lrec-1.497} {{U}niversal
  {D}ependencies v2: An evergrowing multilingual treebank collection}.
\newblock In \emph{Proceedings of the 12th Language Resources and Evaluation
  Conference}, pages 4034--4043, Marseille, France. European Language Resources
  Association.

\bibitem[{Oepen et~al.(2016)Oepen, Kuhlmann, Miyao, Zeman, Cinkov{\'a},
  Flickinger, Haji{\v{c}}, Ivanova, and Ure{\v{s}}ov{\'a}}]{oepen2016towards}
Stephan Oepen, Marco Kuhlmann, Yusuke Miyao, Daniel Zeman, Silvie Cinkov{\'a},
  Dan Flickinger, Jan Haji{\v{c}}, Angelina Ivanova, and Zde{\v{n}}ka
  Ure{\v{s}}ov{\'a}. 2016.
\newblock \href {https://www.aclweb.org/anthology/L16-1630} {Towards
  comparability of linguistic graph {B}anks for semantic parsing}.
\newblock In \emph{Proceedings of the Tenth International Conference on
  Language Resources and Evaluation ({LREC}'16)}, pages 3991--3995,
  Portoro{\v{z}}, Slovenia. European Language Resources Association (ELRA).

\bibitem[{{\v{Z}}abokrtsk{\'y} et~al.(2020){\v{Z}}abokrtsk{\'y}, Zeman, and
  {\v{S}}ev{\v{c}}{\'\i}kov{\'a}}]{zabokrtsky-etal-2020-sentence}
Zden{\v{e}}k {\v{Z}}abokrtsk{\'y}, Daniel Zeman, and Magda
  {\v{S}}ev{\v{c}}{\'\i}kov{\'a}. 2020.
\newblock \href {https://doi.org/10.1162/coli_a_00385} {Sentence meaning
  representations across languages: What can we learn from existing
  frameworks?}
\newblock \emph{Computational Linguistics}, 46(3):605--665.

\end{thebibliography}
\bibliographystyle{acl_natbib}

\end{document}